\pdfoutput=1

\documentclass[11pt]{article}

\usepackage{acl}

\usepackage{times}
\usepackage{latexsym}

\usepackage[T1]{fontenc}

\usepackage[utf8]{inputenc}

\usepackage{microtype}

\usepackage{colortbl}
\usepackage{adjustbox}
\usepackage{amsmath} 
\usepackage{xcolor}
\usepackage{color}
\usepackage{amsfonts} 

\usepackage{subfiles}

\title{AbductionRules: Training Transformers to Explain Unexpected Inputs}

\author{Nathan Young, Qiming Bao, Joshua Bensemann, Michael Witbrock \\
Strong AI Lab\\
School of Computer Science\\
University of Auckland\\
\texttt{\{nathan.young, josh.bensemann, m.witbrock\}@auckland.ac.nz }\\
\texttt{qbao775@aucklanduni.ac.nz}\\

}

\begin{document}

\maketitle

\begin{abstract}
Transformers have recently been shown to be capable of reliably performing logical reasoning over facts and rules expressed in natural language, but abductive reasoning - inference to the best explanation of an unexpected observation - has been underexplored despite significant applications to scientific discovery, common-sense reasoning, and model interpretability.

We present AbductionRules, a group of natural language datasets designed to train and test generalisable abduction over natural-language knowledge bases.
We use these datasets to finetune pretrained Transformers and discuss their performance, finding that our models learned generalisable abductive techniques but also learned to exploit the structure of our data.
Finally, we discuss the viability of this approach to abductive reasoning and ways in which it may be improved in future work.
\end{abstract}

\section{Introduction}

Since its introduction, models based on the Transformer \citep{vaswani2017attention} have, due to their learning ability and Turing-completeness \citep{bhattamishra2020computational}, sparked research into their use in many applications beyond their original purpose of natural language processing (NLP), including image processing and generation \citep{parmar2018image, chen2020generative}, theorem proving \citep{polu2020generative, welleck2021naturalproofs}, and chess \citep{noever2020chess}.

One such task is logical inference - reasoning over first-order logic (FOL) knowledge bases (collections of facts and rules). 
Given a knowledge base, one may attempt to find logical implications (deduction), discover rules that extrapolate patterns in known facts (induction), or infer facts that would explain surprising observations (abduction).
More specifically, if a newly observed fact $p$ cannot be deduced from an existing knowledge base, abduction is the process of finding one or more facts that, if added to the knowledge base, would allow $p$ to be deduced from existing rules.
Figure \ref{fig:explanation} demonstrates the difference between these three kinds of inference.

\begin{figure*}
\centering
\begin{adjustbox}{max width=\textwidth}
\begin{tabular}{l c c c c c}
Deduction:&Socrates is human&$\rightarrow$&Humans are mortal&$\rightarrow$&?\\
\hline
Induction:&Socrates is human&$\rightarrow$&?&$\rightarrow$&Socrates is mortal\\
\hline
Abduction:&?&$\rightarrow$&Humans are mortal&$\rightarrow$&Socrates is mortal\\
\end{tabular}
\end{adjustbox}

\caption{A comparison of deduction, induction, and abduction, as attempts to reconstruct different parts of the same line of FOL reasoning. Note that only deduction is fully reliable, induction may go in either direction in this case, and only abduction produces new knowledge.}
\label{fig:explanation}
\end{figure*}

Traditionally, FOL is represented using a formal mathematical syntax, with facts resembling $\textsc{Human(Socrates)}$ and rules resembling $\forall \textsc{X}: \textsc{Human(X)} \implies \textsc{Mortal(X)}$.
\citet{ruletaker} recently pioneered an alternative approach we call \textit{natural-language logic}, which might represent these as "Socrates is human" and "Humans are mortal". This approach, properly followed, retains the precision of the mathematical syntax while also taking advantage of Transformers' NLP aptitude and pretraining.
This approach also allows reasoning over texts not written in formal representations.

\citet{ruletaker} examined their models' potential for deduction only. \citet{tafjord-etal-2021-proofwriter} extended this work to explore abduction but retained a focus on deduction.

Our goal is to train Transformers to perform abductive reasoning with the following properties:
\begin{itemize}
\item
\textbf{Natural}: Operate over natural language.
\item
\textbf{Generalisable}: Be able to apply techniques outside domains in which they were learned.
\item
\textbf{Generative}: Produce explanations rather than labelling them as sufficient or insufficient.
\item
\textbf{Single-hop}: Produce direct explanations. Instead of "plants are green because chlorophyll is green because green light is not used in photosynthesis", prefer "plants are green because chlorophyll is green". If further explanation is desired, abduction can be applied again.
\item
\textbf{Discerning}: Prefer simpler explanations.
\item
\textbf{Explicit}: Use given knowledge bases rather than relying on pretraining.
\end{itemize}

Our efforts to train abduction in this way are motivated by multiple potential applications.
\begin{itemize}
\item
\citet{Ray2007} describes the use of automated abduction in scientific discovery. Since much scientific knowledge exists in the form of natural language rather than formal representations, advances in natural-language abduction would greatly assist in automating the scientific method by helping to explain experimental observations.
\item
\citet{Ignatiev2018} describe the use of abduction to interpret deep learning models similar to Transformers, which are infamously difficult to interpret.
\item
Abduction may also help solve the longstanding problem of automating common-sense reasoning. Transformers excel at memorising common knowledge but routinely fail to capture any underlying reasoning. 
Training these models to explain their own outputs may remedy this problem by providing a way to integrate this fractured knowledge into a more connected model of reality. 
\end{itemize}

We present the following contributions:
\begin{itemize}
\item
A collection of datasets for training and testing natural-language abduction.
\item
A method of synthetically generating more realistic natural-language logic datasets.
\item
Experimental results showing that Transformers can perform abductive reasoning without additional architecture.
\end{itemize}

\section{Related Work}

\subsection{Natural-language logic}
Our work builds on the RuleTaker line of research on natural-language logic. This line began with \citet{ruletaker}, who developed RuleTakers to reason deductively over FOL knowledge bases expressed in natural language, judging given facts to be true or false. These achieved promising results but failed to accurately explain their reasoning or generalise to inferences requiring more steps than were seen at training time. PRover \cite{prover} achieved greater explainability by generating proofs of its answers. Similarly, the Iterative variant of ProofWriter \citep{tafjord-etal-2021-proofwriter} chained single-hop deductions rather than reasoning through multi-hop deductions all at once, making its reasoning transparent and easily generalisable to unseen depths.
multiPRover \citep{saha-etal-2021-multiprover} also made use of this iterative approach.
The generalisability and interpretability of iterative single-hop reasoning are why we seek to train single-hop abduction.

\citet{tafjord-etal-2021-proofwriter} also adapted their deduction-based datasets to train abductive reasoning, achieving success but training multi-hop abduction only, and also requiring models to output every possible explanation.
By contrast, we seek to train models to discern between simpler and more complex explanations - for example, to prefer explanations requiring fewer unknown facts.

\subsection{Other adjacent work}
\citet{bhagavatula2019abductive} presented two more abduction-based datasets: $\alpha$-NLI, which tests models' ability to choose which of two hypotheses better explains an observation, and $\alpha$-NLG, a generative version of the same dataset.
These datasets do not give supporting knowledge bases - all background information must come from pretraining. While this is a valuable approach, we seek to investigate how well Transformers can reason over given knowledge bases to incorporate explicit background knowledge.

\citet{gontier2020measuring} investigated Transformers' ability to perform inductive reasoning in natural language, finding them able to extrapolate patterns in given proofs but again unable to generalise to more complex proofs.

\citet{saparov2021generative} developed an alternative approach to classifying the ProofWriter datasets that does not reason over natural language, instead using a symbolic, Bayesian approach and using abductive reasoning to satisfy constraints. Their models' superior performance demonstrates that while Transformers are effective at logical reasoning, they may benefit from more specialised architecture.

\section{Methodology}
Prior to our work, there existed no dataset capable of training or testing the kind of abductive reasoning we seek.
We therefore present \textbf{AbductionRules}, a natural-language logic dataset designed for this task, and use it to train and test several models based on a pretrained Text-to-Text Transfer Transformer, or T5 \citep{t5}.

\subsection{Datasets}

AbductionRules has three main predecessors.

\subsubsection{Rule Reasoning}
The Rule Reasoning dataset developed by \citet{ruletaker} was, to our knowledge, the first natural-language logic dataset.

To create this dataset, FOL predicates (e.g. \textsc{Big(Lion)}) were procedurally generated, entities (\textsc{Lion}) and attributes (\textsc{Big(X)}) were extracted, and templates ("The \{entity\} is \{attribute\}") were used to create natural-language logic translations ("The lion is big"). Rules were created similarly (e.g. $\forall \textsc{X}: \textsc{Big(X)} \implies \textsc{Blue(X)}$ became "If something is big then it is blue").
Facts and rules were grouped into knowledge bases, each with several questions; the model's task is to label each question true or false.

The Rule Reasoning dataset includes knowledge bases in several domains; those in the \textit{animal-domain} use animals as entities while those in the \textit{person-domain} use peoples' names. All subsequent datasets similarly use these two domains.
The animal-domain includes \textit{multi-entity} facts (\textsc{Chases(Lion, Mouse)}, or "the lion chases the mouse"). For our purposes, we consider the lion to be the main entity and "chases the mouse" to be an attribute of the lion.

\subsubsection{ParaRules}
Recognising that their translations of mathematical syntax into natural language were strict and unrealistic (e.g. "Charlie is green. Charlie is rough."), \citet{ruletaker} also produced ParaRules, which contained knowledge bases and questions similar to those in the Rule Reasoning dataset, but were paraphrased into more colloquial language (e.g. "Charlie has green teeth and rough skin."). This approach much better prepares Transformers to reason logically over naturally-occurring texts but requires large amounts of human labour to produce. For this reason, ParaRules is much smaller than the Rule Reasoning dataset.

\subsubsection{PARARULE Plus}
Seeing the value in RuleTaker's size and ease of production as well as the greater utility of ParaRules, \citet{PARARULEPlus} produced PARARULE Plus, a compromise between the Rule Reasoning dataset and ParaRules that procedurally rephrases all rules during generation by using various templates. PARARULE Plus also avoids eschewingz word associations entirely by grouping related attributes (such as "big", "strong", "high" and "huge") and only giving entities attributes from one group. While PARARULE Plus falls short of ParaRules' variety, its greater collection of rephrased rules is highly valuable.

\subsubsection{AbductionRules}

We adapt the open-source code used to generate PARARULE Plus to create AbductionRules\footnote{AbductionRules can be found and downloaded at \url{https://github.com/Strong-AI-Lab/AbductionRules/tree/main/datasets/}.}, making the following changes:

\begin{itemize}
\item
Instead of labelling questions (for our purposes, "observations") with "true" or "false", we use the lone fact (or "explanation") that would prove or disprove it.
\item
We ensure that no two knowledge bases in the same dataset give the same attributes to the same entities to avoid repeats. This reduces the size of the datasets; to compensate, we increase the number of entities.
\item
While each rule has a single condition in PARARULE Plus ("If something is cute, then..."), we give three ("If something is cute, funny, and adorable, then..."), with an entity that satisfies exactly two conditions; the model must identify the third.
\end{itemize}

\begin{figure}[h]

\begin{adjustbox}{max width=\columnwidth}
\fbox{
 \parbox{\columnwidth}{
 
\textit{Context(Facts+Rules):}

\textit{Facts:}
\textbf{The squirrel is \colorbox{cyan}{quiet}.}
The leopard is slow.
The dog is adorable.
The crocodile is heavy.
The leopard is boring.
The leopard is angry.
The crocodile is awful.
The leopard attacks the squirrel.
The dog is small.
The dog is cute.
The squirrel is nice.
The crocodile likes the dog.
\textbf{The squirrel is \colorbox{pink}{kind}.}

\textit{Rules:}
If something is cute, is adorable, and is furry, then it is also lovely.
All animals that are obese, are awful, and are heavy, are big.
If an animal is fierce, sees the squirrel, and likes the dog, it is tired.
\textbf{Things that are \colorbox{green}{smart}, are \colorbox{pink}{kind}, and are \colorbox{cyan}{quiet}, are also \colorbox{yellow}{round}.}
If an animal chases the dog, is boring, and attacks the squirrel, then it is also strong.
All things that are slow, are sleepy, and are angry, are rough. 

\textit{Observation:} The squirrel is \colorbox{yellow}{round}.

\textit{Explanation:} The squirrel is \colorbox{green}{smart}.

 }
}
\end{adjustbox}
\caption{An example observation, explanation, and corresponding context from Abduction-Animal-Simple.
The model must output the explanation given the context and observation as input.
Facts and rules used to explain the observation are bolded while relevant attributes are highlighted.
}	
\label{animalsimpleexample}
\end{figure}

After making these changes, we produce datasets with increasing levels of complexity.

\begin{figure*}[t]
\begin{adjustbox}{max width=\textwidth}
\begin{tabular}{|l|l|l|l|}
\hline
Initial & Shuffled & Rephrased & Confounded \\
\hline
\colorbox{red!20}{The cat is round.} & \colorbox{orange!20}{The cat is smart.} & \colorbox{orange!20}{The cat is smart.} & \colorbox{orange!20}{The cat is smart.} \\
\colorbox{orange!20}{The cat is smart.} & \colorbox{yellow!20}{If something is round,} & \colorbox{lime!20}{All animals that are} & \colorbox{lime!20}{All animals that are} \\
\colorbox{yellow!20}{If something is round,} & \colorbox{yellow!20}{smart, and quiet, then} & \colorbox{lime!20}{round, are smart, and} & \colorbox{lime!20}{round, are smart, and} \\
\colorbox{yellow!20}{smart, and quiet, then} & \colorbox{yellow!20}{it is kind.} & \colorbox{lime!20}{are quiet, are also kind.} & \colorbox{lime!20}{are quiet, are also kind.} \\
\colorbox{yellow!20}{it is kind.} & \colorbox{red!20}{The cat is round.} & \colorbox{red!20}{The cat is round.} & \colorbox{red!20}{The cat is round.} \\
 & & & \colorbox{green!20}{If an animal is round,} \\
 & & & \colorbox{green!20}{is boring, and is quiet,} \\
 & & & \colorbox{green!20}{it is kind.} \\
\hline
\end{tabular}
\end{adjustbox}
\caption{\label{tab:complexity}A diagram demonstrating the successive changes we make to the AbductionRules knowledge bases.}
\end{figure*}

\begin{itemize}
\item
The first complexity level contains no further changes from PARARULE Plus and yields the dataset \textbf{Abduction-Animal-0.1}.

\item
At the second complexity level, we shuffle all knowledge bases to prevent models from exploiting the constant position of all sentences and attributes. This yields the dataset \textbf{Abduction-Animal-0.2}.

\item
At the third complexity level, we procedurally rephrase rules with random variations instead of using the same templates as PARARULE Plus. For example, the animal-domain FOL rule $\forall \textsc{X}: (\textsc{Big(X)} \land \textsc{Heavy(X)} \land \textsc{Fierce(X))} \implies \textsc{Strong(X)}$ might be rephrased as "All animals that are big, are heavy, and are fierce, are also strong" or "If something is heavy, is fierce, and is big, it is strong", among many other similar variations. Notably, this rephrasing process involves reordering all attributes so that attributes contained in correct abductions might be first, second, or third. This yields the datasets \textbf{Abduction-Animal-Simple} and \textbf{Abduction-Person-Simple.}
\\
Figure \ref{animalsimpleexample} contains an example item from Animal-Simple.\footnote{For brevity, we omit the "Abduction-" prefix when discussing the AbductionRules datasets within this paper.}

This method of procedural rule rephrasing represents a useful iteration on the natural-language logic approach and leaves room for further improvement.
Concentrated work in this line of research may produce synthetic natural-language logic datasets that are larger yet exhibit much wider variety, making this approach more powerful and robust.

\item
At the fourth and final complexity level, we add extraneous confounding rules to knowledge bases. While lower complexity levels only ever have one rule that could explain a given observation, here we create two variations of every (single-entity) rule; one replaces a satisfied condition with an unsatisfied condition, while the other replaces all three conditions. 
All replacements come from different pools. 
This yields the datasets \textbf{Abduction-Animal} and \textbf{Abduction-Person}.
\end{itemize}

Figure \ref{tab:complexity} contains simplified examples of data from each complexity level.

We intend each successive complexity level to remove additional idiosyncrasies that might be exploited in lieu of using abduction (i.e. used to "cheat"), so that this exploitation can be detected. 
We also intend the fourth to train models to favour simpler explanations when strictly more complex explanations are available.

\subsection{Experiments}
We use AbductionRules to train 8 models based on the pretrained T5 implementation from the HuggingFace Transformers library \citep{wolf-etal-2020-transformers}.\footnote{All code used for experiments in this paper can be found at \url{https://github.com/Strong-AI-Lab/AbductionRules/}.}

We first use each training set to train 1 model, yielding 6 models trained at 4 complexity levels across 2 domains.
To compare domains and complexity levels, we test all models on all test sets, giving us intra-domain results (isolating the effect of the complexity), and inter-domain results (some isolating the effect of the domain).
We expect each successive complexity level to train a better-quality model and the two domains to be mostly comparable with some variation attributable to the animal-domain's multi-entity facts.

If our approach were adapted to models extensively trained to reason on many domains, we expect that teaching abduction in every domain would be prohibitively expensive. Therefore, we seek to investigate Transformers' ability to transfer abductive reasoning techniques to domains where these techniques have not been taught but are nonetheless familiar to the Transformer. To this end, we train two more multi-domain models.
\begin{itemize}
\item
We train one model on our simplest dataset and our most complex dataset in another domain, i.e. Animal-0.1 and Person. We name this model \textit{Person+Animal-0.1}.
\item
We train another model on the simplest person-domain dataset and the most complex animal-domain dataset, i.e. Person-Simple and Animal, to compare the two domains. We name this model \textit{Animal+Person-Simple}.

\end{itemize}


While we are interested in these multi-domain models' performance on all datasets, we are particularly interested in their results on the most complex dataset on which they were not trained (Abduction-Animal and Abduction-Person, respectively). We treat performance on these datasets as a proxy for Transformers' ability to apply abductive reasoning outside the domains in which it was trained.

Finally, we use the existing pretrained T5 model with no additional training as our baseline model.

\section{Results}

\begin{table*}[t]
\begin{adjustbox}{max width=\textwidth}
\begin{tabular}{|c|c|c|c|c|c|c|}
\hline
Model $\backslash$ Test set&Animal-0.1&Animal-0.2&Animal-Simple&Animal&Person-Simple&Person\\
\hline
Untrained&\cellcolor{blue!0.0}0.0\%&\cellcolor{blue!0.0}0.0\%&\cellcolor{blue!0.0}0.0\%&\cellcolor{blue!0.0}0.0\%&\cellcolor{blue!0.0}0.0\%&\cellcolor{blue!0.0}0.0\%\\
\hline
Abduction-Animal-0.1&\cellcolor{blue!50.0}\textbf{100.0\%}&\cellcolor{blue!49.65}99.3\%&\cellcolor{blue!24.0}48.0\%&\cellcolor{blue!14.4}28.8\%&\cellcolor{blue!0.0}0.0\%&\cellcolor{blue!0.0}0.0\%\\
\hline
Abduction-Animal-0.2&\cellcolor{blue!50.0}100.0\%&\cellcolor{blue!50.0}\textbf{100.0\%}&\cellcolor{blue!18.85}37.7\%&\cellcolor{blue!11.6}23.2\%&\cellcolor{blue!0.0}0.0\%&\cellcolor{blue!0.0}0.0\%\\
\hline
Abduction-Animal-Simple&\cellcolor{blue!50.0}100.0\%&\cellcolor{blue!50.0}100.0\%&\cellcolor{blue!50.0}\textbf{100.0\%}&\cellcolor{blue!25.05}50.1\%&\cellcolor{blue!0.0}0.0\%&\cellcolor{blue!0.0}0.0\%\\
\hline
Abduction-Animal&\cellcolor{blue!46.3}92.6\%&\cellcolor{blue!46.75}93.5\%&\cellcolor{blue!47.05}94.1\%&\cellcolor{blue!50.0}\textbf{100.0\%}&\cellcolor{blue!0.0}0.0\%&\cellcolor{blue!0.0}0.0\%\\
\hline
Abduction-Person-Simple&\cellcolor{blue!0.0}0.0\%&\cellcolor{blue!0.0}0.0\%&\cellcolor{blue!0.0}0.0\%&\cellcolor{blue!0.0}0.0\%&\cellcolor{blue!50.0}\textbf{100.0\%}&\cellcolor{blue!12.8}25.6\%\\
\hline
Abduction-Person&\cellcolor{blue!0.0}0.0\%&\cellcolor{blue!0.0}0.0\%&\cellcolor{blue!0.0}0.0\%&\cellcolor{blue!0.0}0.0\%&\cellcolor{blue!13.4}26.8\%&\cellcolor{blue!50.0}\textbf{100.0\%}\\
\hline
Person+Animal-0.1&\cellcolor{blue!50.0}\textbf{100.0\%}&\cellcolor{blue!50.0}100.0\%&\cellcolor{blue!38.35}76.7\%&\cellcolor{blue!42.75}85.5\%&\cellcolor{blue!46.45}92.9\%&\cellcolor{blue!50.0}\textbf{100.0\%}\\
\hline
Animal+Person-Simple&\cellcolor{blue!49.55}99.1\%&\cellcolor{blue!49.55}99.1\%&\cellcolor{blue!49.7}99.4\%&\cellcolor{blue!50.0}\textbf{100.0\%}&\cellcolor{blue!50.0}\textbf{100.0\%}&\cellcolor{blue!49.9}99.8\%\\
\hline
\end{tabular}
\end{adjustbox}
\caption{\label{tab:correctresults}Performance of all models on all test sets. Test sets corresponding to training sets are bolded.}
\end{table*}

Table \ref{tab:correctresults} contains our results, showing the percentage of abductions correctly performed by each model on each test set.
\footnote{All our results can be found at \url{https://github.com/Strong-AI-Lab/AbductionRules/tree/main/results/}.}

Note that no model ever gave a correct answer in a domain on which it was not trained.
On the surface, this would suggest that our models were unable to generalise to new domains.
However, inspection of inter-domain results shows that this is not entirely accurate; many explanations contain errors but nonetheless identify the ground-truth explanation. 
For example, the animal models commonly appended "The" to correct explanations, as in "The Bob is small"; while this is incorrect, it nonetheless indicates the correct explanation in a way that suggests the model still performed the correct abduction.
We distinguish between two kinds of errors in correct-yet-useful explanations: \textit{lossless errors} and \textit{lossy errors}.

\subsection{Lossless errors}
Explanations with lossless errors failed to match the correct explanation character-for-character but allowed it to be reliably identified.

\begin{table*}[t]
\begin{adjustbox}{max width=\textwidth}
\begin{tabular}{|c|c|c|c|c|c|c|}
\hline
Model $\backslash$ Test set&Animal-0.1&Animal-0.2&Animal-Simple&Animal&Person-Simple&Person\\
\hline
Untrained&\cellcolor{blue!0.0}0.0\%&\cellcolor{blue!0.0}0.0\%&\cellcolor{blue!0.0}0.0\%&\cellcolor{blue!0.0}0.0\%&\cellcolor{blue!0.0}0.0\%&\cellcolor{blue!0.0}0.0\%\\
\hline
Abduction-Animal-0.1&\cellcolor{blue!50.0}\textbf{100.0\% (-)}&\cellcolor{blue!49.65}99.3\% (-)&\cellcolor{blue!24.0}48.0\% (-)&\cellcolor{blue!14.4}28.8\% (-)&\cellcolor{blue!6.6}13.2\% (+13.2\%)&\cellcolor{blue!5.05}10.1\% (+10.1\%)\\
\hline
Abduction-Animal-0.2&\cellcolor{blue!50.0}100.0\% (-)&\cellcolor{blue!50.0}\textbf{100.0\% (-)}&\cellcolor{blue!19.25}38.5\% (+0.9\%)&\cellcolor{blue!11.8}23.6\% (+0.4\%)&\cellcolor{blue!4.8}9.6\% (+9.6\%)&\cellcolor{blue!2.85}5.7\% (+5.7\%)\\
\hline
Abduction-Animal-Simple&\cellcolor{blue!50.0}100.0\% (-)&\cellcolor{blue!50.0}100.0\% (-)&\cellcolor{blue!50.0}\textbf{100.0\% (-)}&\cellcolor{blue!25.05}50.1\% (-)&\cellcolor{blue!17.2}34.4\% (+34.4\%)&\cellcolor{blue!3.5}7.0\% (+7.0\%)\\
\hline
Abduction-Animal&\cellcolor{blue!46.3}92.6\% (-)&\cellcolor{blue!46.75}93.5\% (-)&\cellcolor{blue!47.1}94.2\% (+0.0\%)&\cellcolor{blue!50.0}\textbf{100.0\% (-)}&\cellcolor{blue!12.5}25.0\% (+25.0\%)&\cellcolor{blue!18.25}36.5\% (+36.5\%)\\
\hline
Abduction-Person-Simple&\cellcolor{blue!0.75}1.5\% (+1.5\%)&\cellcolor{blue!0.65}1.3\% (+1.3\%)&\cellcolor{blue!0.45}0.9\% (+0.9\%)&\cellcolor{blue!0.15}0.3\% (+0.3\%)&\cellcolor{blue!50.0}\textbf{100.0\% (-)}&\cellcolor{blue!12.8}25.6\% (-)\\
\hline
Abduction-Person&\cellcolor{blue!0.0}0.0\% (-)&\cellcolor{blue!0.0}0.0\% (-)&\cellcolor{blue!0.0}0.0\% (-)&\cellcolor{blue!0.0}0.0\% (-)&\cellcolor{blue!13.4}26.8\% (-)&\cellcolor{blue!50.0}\textbf{100.0\% (-)}\\
\hline
Person+Animal-0.1&\cellcolor{blue!50.0}\textbf{100.0\% (-)}&\cellcolor{blue!50.0}100.0\% (-)&\cellcolor{blue!38.35}76.7\% (-)&\cellcolor{blue!42.75}85.5\% (-)&\cellcolor{blue!46.45}92.9\% (-)&\cellcolor{blue!50.0}\textbf{100.0\% (-)}\\
\hline
Animal+Person-Simple&\cellcolor{blue!49.55}99.1\% (-)&\cellcolor{blue!49.55}99.1\% (-)&\cellcolor{blue!49.7}99.4\% (-)&\cellcolor{blue!50.0}\textbf{100.0\% (-)}&\cellcolor{blue!50.0}\textbf{100.0\% (-)}&\cellcolor{blue!49.9}99.8\% (-)\\
\hline
\end{tabular}
\end{adjustbox}
\caption{\label{tab:fixableresults}Improvement of all models on all test sets after allowing lossless errors.}
\end{table*}

We found several ways in which recognisably correct explanations differed from the ground-truth, such as extra words ("The Bob is small", "The lion is attacks the mouse"), looping ("The dog is is is is is small"), and incorrect grammar ("The anne is wealthy").
While these errors point towards flaws in training, it is a strength of natural-language logic and soft reasoners that they can cope with minor grammar mistakes as long as meaning is preserved.

Table \ref{tab:fixableresults} contains our results after correcting for these errors.
Note that animal-domain models achieved performance comparable to the person-domain models on novel datasets in their own domain, while person-domain models saw minimal inter-domain improvement.
multi-domain models also saw almost no improvement, suggesting that having seen correct explanations in both domains eliminated this kind of formatting error.

\subsection{Lossy errors}
The most important aspect of abduction in our datasets is identification of the correct attribute. The entity at the beginning of the explanation always matches that at the beginning of the observation; therefore, if the correct attribute is identified, the correct explanation can be reconstructed.


\begin{table*}[t]
\begin{adjustbox}{max width=\textwidth}
\begin{tabular}{|c|c|c|c|c|c|c|}
\hline
Model $\backslash$ Test set&Animal-0.1&Animal-0.2&Animal-Simple&Animal&Person-Simple&Person\\
\hline
Untrained&\cellcolor{blue!0.0}0.0\%&\cellcolor{blue!0.1}0.2\%&\cellcolor{blue!0.05}0.1\%&\cellcolor{blue!0.0}0.0\%&\cellcolor{blue!0.0}0.0\%&\cellcolor{blue!0.0}0.0\%\\
\hline
Abduction-Animal-0.1&\cellcolor{blue!50.0}\textbf{100.0\% (-)}&\cellcolor{blue!50.0}100.0\% (+0.7\%)&\cellcolor{blue!24.2}48.4\% (+0.4\%)&\cellcolor{blue!14.55}29.1\% (+0.3\%)&\cellcolor{blue!20.95}41.9\% (+28.8\%)&\cellcolor{blue!11.75}23.5\% (+13.4\%)\\
\hline
Abduction-Animal-0.2&\cellcolor{blue!50.0}100.0\% (-)&\cellcolor{blue!50.0}\textbf{100.0\% (-)}&\cellcolor{blue!19.7}39.4\% (+0.8\%)&\cellcolor{blue!12.25}24.5\% (+0.9\%)&\cellcolor{blue!14.7}29.4\% (+19.8\%)&\cellcolor{blue!7.15}14.3\% (+8.6\%)\\
\hline
Abduction-Animal-Simple&\cellcolor{blue!50.0}100.0\% (-)&\cellcolor{blue!50.0}100.0\% (-)&\cellcolor{blue!50.0}\textbf{100.0\% (-)}&\cellcolor{blue!25.05}50.1\% (-)&\cellcolor{blue!33.2}66.4\% (+32.0\%)&\cellcolor{blue!7.45}14.9\% (+7.9\%)\\
\hline
Abduction-Animal&\cellcolor{blue!46.3}92.6\% (-)&\cellcolor{blue!46.75}93.5\% (-)&\cellcolor{blue!47.1}94.2\% (-)&\cellcolor{blue!50.0}\textbf{100.0\% (-)}&\cellcolor{blue!19.55}39.1\% (+14.1\%)&\cellcolor{blue!31.35}62.7\% (+26.2\%)\\
\hline
Abduction-Person-Simple&\cellcolor{blue!19.9}39.8\% (+38.4\%)&\cellcolor{blue!21.4}42.8\% (+41.5\%)&\cellcolor{blue!19.8}39.6\% (+38.7\%)&\cellcolor{blue!5.75}11.5\% (+11.2\%)&\cellcolor{blue!50.0}\textbf{100.0\% (-)}&\cellcolor{blue!12.8}25.6\% (-)\\
\hline
Abduction-Person&\cellcolor{blue!2.6}5.2\% (+5.2\%)&\cellcolor{blue!2.5}5.0\% (+5.0\%)&\cellcolor{blue!2.45}4.9\% (+4.9\%)&\cellcolor{blue!7.9}15.8\% (+15.8\%)&\cellcolor{blue!13.4}26.8\% (-)&\cellcolor{blue!50.0}\textbf{100.0\% (-)}\\
\hline
Person+Animal-0.1&\cellcolor{blue!50.0}\textbf{100.0\% (-)}&\cellcolor{blue!50.0}100.0\% (-)&\cellcolor{blue!38.35}76.7\% (-)&\cellcolor{blue!42.8}85.6\% (+0.0\%)&\cellcolor{blue!46.45}92.9\% (-)&\cellcolor{blue!50.0}\textbf{100.0\% (-)}\\
\hline
Animal+Person-Simple&\cellcolor{blue!49.55}99.1\% (-)&\cellcolor{blue!49.55}99.1\% (-)&\cellcolor{blue!49.7}99.4\% (-)&\cellcolor{blue!50.0}\textbf{100.0\% (-)}&\cellcolor{blue!50.0}\textbf{100.0\% (-)}&\cellcolor{blue!49.9}99.8\% (-)\\
\hline
\end{tabular}
\end{adjustbox}
\caption{\label{tab:usefulresults}Improvement of all models on all test sets after allowing lossy errors.}
\end{table*}

Table \ref{tab:usefulresults} contains our results after correcting for these errors.
Note that every model achieved some useful results on every test set.
Most inter-domain results improved to rival intra-domain results, although the Abduction-Person model continued to struggle.
Intra-domain results saw minimal improvement, with none seeing a >2\% point increase.
The multi-domain models again saw no visible improvement, further suggesting that these inferior results were avoidable from seeing facts, rules, and explanations in different formats at training time.

\section{Discussion}
Our results show that models trained on our simplest datasets struggle to generalise to new complexity levels and domains, while those trained on our more complex datasets are better able to generalise but still perform suboptimally. Meanwhile, those trained on combined multi-domain datasets achieve performance superior to the sum of models trained on their parts and easily apply skills outside domains in which they were learned.
It is also clear that models trained in the animal-domain achieve better intra-domain and inter-domain performance than person-domain models.

The untrained T5 model performs abysmally and merits little discussion, indicating that this abduction task is non-trivial (at least in the way we present it here).

\subsection{Animal-0.1 and Animal-0.2}
Unsurprisingly, the models trained on our simplest datasets fare the worst of our trained models. Our Animal-0.1 and -0.2 models perform similarly poorly, suggesting that Animal-0.2's additional complexity from randomised sentence orderings was of minimal importance. 
In fact, the Animal-0.2 model's performance on more complex datasets is worse than its simpler counterpart; examination of its results reveals a tendency to loop on unfamiliar inputs. Given the Animal-0.1 model's 99.3\% correct (100\% allowing lossy errors) performance on Animal-0.2, we treat these complexity levels as equivalent and the Animal-0.1 model as definitive.

The Animal-0.1 model is approximately 1/3 as accurate on the person-domain when allowing lossless errors but only loses approximately 6\% points when allowing lossy errors, suggesting that it fails to adapt to new formats but is mostly able to use the same techniques as in the animal-domain.

These models' significant performance hit on higher complexity levels clearly indicates that they exploit the structure of their training set. However, it should be noted that the Animal-0.1 model drops each time by approximately a factor of 2. If this model only chose the penultimate attribute in a sentence containing the attribute in the question, its accuracy would drop by a factor of 3 with procedural rephrasing and again with confounding rules.
Therefore, both models utilise some level of generalised abductive reasoning.

\subsection{Animal-Simple and Person-Simple}
The Animal-Simple model significantly outperforms our simpler models; this makes sense since Animal-0.1 and -0.2 can be thought of as special, unshuffled cases of Animal-Simple.\footnote{Because of this and their failure to train generalisable abduction, we do not include either Animal-0.1 or -0.2 in the public release of AbductionRules. The code we used for our experiments can be used to regenerate them if desired.}
Similarly to the Animal-0.1 model, the Animal-Simple model performs about half as well on Animal as on Animal-Simple. 
This model also performs worse on Person-Simple than Animal when allowing lossless errors but better when allowing lossy errors, implying that it exploits the structure of Animal-Simple to some degree to identify correct attributes.
Its performance drop from Person-Simple to Person is greater than from Animal-Simple to Animal, suggesting that changes in domain and complexity are more difficult to generalise when compounded. 

Our Person-Simple model also performs well but fails to generalise to higher complexity; this can be partially explained by the multi-entity facts in the animal-domain, as rules using these facts are not used to create confounding rules.
This model gives almost no correct inter-domain explanations unless lossy errors are allowed, in which case it achieves similar inter-domain performance to the animal-domain models. Its performance drop on Animal can be compared to that of the Animal-Simple model from Person-Simple to Person, exacerbated by the person-domain models' poorer performance in general.

\subsection{Abduction-Animal and Abduction-Person}
The Animal model performs the best of all single-domain models, achieving >60\% performance on all datasets except Person-Simple when allowing lossy errors. The drop from Person to Person-Simple is evidence of cheating, but its generalisability is superior to all other models and demonstrates some abductive ability.
Surprisingly, it achieves worse intra-domain results on lower complexity levels than the Animal-Simple model, again indicating that some of its performance is dependent on Animal's rule structure. Still, this performance drop is relatively small (being <10\% in all cases), further reinforcing that while this model utilises some degree of both cheating and abduction (like all our models), its abductive capabilities generalise to a promising extent.

By contrast, the Person model achieves the worst performance of any model, performing as well on Person-Simple as that dataset's model does on Person and achieving abysmal inter-domain performance, even on Animal. 
This model is the clearest indication that (our instantiations of) the two domains are not equivalent; the animal-domain's models are much better able to generalise.
The multi-entity rules again offer some explanatory power - the Animal model demonstrates some overtraining on the confounding rules and so performs more poorly in their absence, but still learned to explain observations using multi-entity rules that lacked confounding equivalents, making it robust to extraneous rules but not reliant on them. If this were a major determining factor, we would expect models trained on both maximally and minimally complex datasets to be even more robust and generalised.


\subsection{Multi-domain models}
Our multi-domain models are our best-performing models by far, achieving superior performance on unseen datasets than the sum of models trained on their combined training sets' parts.

The Person+Animal-0.1 model, being trained on our simplest dataset and having its most complex training set come from the worse of our two training domains, is the worse of our two multi-domain models. Nonetheless, it reaches a remarkable level of performance, explaining >76\% of all observations correctly on all test sets.
Its performance in the face of unconfounded rephrased rules (something unprecedented in its training) is dependent on the domain. In the person-domain (i.e. on Person-Simple), where it received its most complex training, it achieves its best result on a dataset it was not trained on (excepting Animal-0.2), while in the animal-domain (i.e. on Animal-Simple) it achieves its worst result, having not seen any rephrased animal rules at training time. Still, it demonstrates a greater ability than any single-domain model to generalise to these unfamiliar rule structures.
It can also apply its training on confounded rules outside the domain in which it was learned, achieving far greater performance on Animal than any other dataset that it was not trained on.

The Animal+Person-Simple model is our best and most promising, achieving >99\% performance on every dataset and consistently adapting to all complexity levels in every domain.
Like Person+Animal-0.1, it encounters unprecedented rule structures (singular single-entity animal rules, confounded person rules) and generalises almost perfectly to each.
While our datasets remain somewhat limited in scope, we believe that this result demonstrates that Transformers can generalise abductive techniques beyond the domains in which those techniques were trained, provided the domain itself is not entirely novel.

Extrapolating these multi-domain results, it seems likely that finetuning Transformers that have received extensive pretraining (such as GPT-3 \citep{brown2020language}) on datasets covering more varied and complex examples of abduction would make these models capable of much more generalised natural-language abductive reasoning.

\section{Conclusion}
We have presented the AbductionRules datasets and shown that pretrained T5 models finetuned on them exhibit generalised abductive reasoning. Our more complex datasets train abduction more generally and reliably than our less complex datasets. Further, training in multiple domains is superior to training in only one domain, and we have clear evidence of generalisation of techniques from one domain to another.

We have presented an innovation in natural-language logic dataset generation, presenting a new middle-ground between the template-based PARARULE Plus \citep{PARARULEPlus} and the manually rephrased Pararules \cite{ruletaker}.
We believe our results are promising and demonstrate the viability of Transformer-based abduction (and logical reasoning in general), but also indicate opportunities for improvement.

\subsection{Future Work}
Future work in this area might explore:
\begin{itemize}
\item
Examining skill transfer between different kinds of logical reasoning.
\item
Applying abductive techniques in real-world, as opposed to artificial, domains.
\item
Generating probability distributions over multiple possible explanations.
\item
Testing explanations by verifying that they allow the original observation to be deduced.
\item
Explanations that include not only missing premises but the relevant rule(s) they satisfy.
\end{itemize}


\bibliography{anthology,custom}

\begin{thebibliography}{20}
\expandafter\ifx\csname natexlab\endcsname\relax\def\natexlab#1{#1}\fi

\bibitem[{Bao(2021)}]{PARARULEPlus}
Qiming Bao. 2021.
\newblock \href {https://github.com/Strong-AI-Lab/PARARULE-Plus} {Pararule
  plus: A larger deep multi-step reasoning dataset over natural language}.

\bibitem[{Bhagavatula et~al.(2019)Bhagavatula, Le~Bras, Malaviya, Sakaguchi,
  Holtzman, Rashkin, Downey, Yih, and Choi}]{bhagavatula2019abductive}
Chandra Bhagavatula, Ronan Le~Bras, Chaitanya Malaviya, Keisuke Sakaguchi, Ari
  Holtzman, Hannah Rashkin, Doug Downey, Wen-tau Yih, and Yejin Choi. 2019.
\newblock \href {https://doi.org/10.48550/arXiv.1908.05739} {Abductive
  commonsense reasoning}.
\newblock In \emph{International Conference on Learning Representations}.

\bibitem[{Bhattamishra et~al.(2020)Bhattamishra, Patel, and
  Goyal}]{bhattamishra2020computational}
Satwik Bhattamishra, Arkil Patel, and Navin Goyal. 2020.
\newblock \href {https://dx.doi.org/10.18653/v1/2020.conll-1.37} {On the
  computational power of transformers and its implications in sequence
  modeling}.
\newblock In \emph{Proceedings of the 24th Conference on Computational Natural
  Language Learning}, pages 455--475.

\bibitem[{Brown et~al.(2020)Brown, Mann, Ryder, Subbiah, Kaplan, Dhariwal,
  Neelakantan, Shyam, Sastry, Askell et~al.}]{brown2020language}
Tom~B Brown, Benjamin Mann, Nick Ryder, Melanie Subbiah, Jared Kaplan, Prafulla
  Dhariwal, Arvind Neelakantan, Pranav Shyam, Girish Sastry, Amanda Askell,
  et~al. 2020.
\newblock \href {https://doi.org/10.48550/arXiv.2005.14165} {Language models
  are few-shot learners}.
\newblock \emph{arXiv preprint arXiv:2005.14165}.

\bibitem[{Chen et~al.(2020)Chen, Radford, Child, Wu, Jun, Luan, and
  Sutskever}]{chen2020generative}
Mark Chen, Alec Radford, Rewon Child, Jeffrey Wu, Heewoo Jun, David Luan, and
  Ilya Sutskever. 2020.
\newblock \href {http://proceedings.mlr.press/v119/chen20s} {Generative
  pretraining from pixels}.
\newblock In \emph{International Conference on Machine Learning}, pages
  1691--1703. PMLR.

\bibitem[{Clark et~al.(2020)Clark, Tafjord, and Richardson}]{ruletaker}
Peter Clark, Oyvind Tafjord, and Kyle Richardson. 2020.
\newblock \href {https://doi.org/10.24963/ijcai.2020/537} {Transformers as soft
  reasoners over language}.
\newblock In \emph{Proceedings of the Twenty-Ninth International Joint
  Conference on Artificial Intelligence (IJCAI-20)}, pages 3882--3890.

\bibitem[{Gontier et~al.(2020)Gontier, Sinha, Reddy, and
  Pal}]{gontier2020measuring}
Nicolas Gontier, Koustuv Sinha, Siva Reddy, and Chris Pal. 2020.
\newblock \href {https://doi.org/10.48550/arXiv.2009.14786} {Measuring
  systematic generalization in neural proof generation with transformers}.
\newblock \emph{Advances in Neural Information Processing Systems}, 33.

\bibitem[{Ignatiev et~al.(2019)Ignatiev, Narodytska, and
  Marques-Silva}]{Ignatiev2018}
Alexey Ignatiev, Nina Narodytska, and Joao Marques-Silva. 2019.
\newblock \href {https://doi.org/10.1609/aaai.v33i01.33011511} {Abduction-based
  explanations for machine learning models}.
\newblock In \emph{Proceedings of the AAAI Conference on Artificial
  Intelligence}, volume~33, pages 1511--1519.

\bibitem[{Noever et~al.(2020)Noever, Ciolino, and Kalin}]{noever2020chess}
David Noever, Matt Ciolino, and Josh Kalin. 2020.
\newblock \href {https://doi.org/10.48550/arXiv.2008.04057} {The chess
  transformer: Mastering play using generative language models}.
\newblock \emph{arXiv preprint arXiv:2008.04057}.

\bibitem[{Parmar et~al.(2018)Parmar, Vaswani, Uszkoreit, Kaiser, Shazeer, Ku,
  and Tran}]{parmar2018image}
Niki Parmar, Ashish Vaswani, Jakob Uszkoreit, Lukasz Kaiser, Noam Shazeer,
  Alexander Ku, and Dustin Tran. 2018.
\newblock \href {https://doi.org/10.48550/arXiv.1802.05751} {Image
  transformer}.
\newblock In \emph{International Conference on Machine Learning}, pages
  4055--4064. PMLR.

\bibitem[{Polu and Sutskever(2020)}]{polu2020generative}
Stanislas Polu and Ilya Sutskever. 2020.
\newblock \href {https://doi.org/10.48550/arXiv.2009.03393} {Generative
  language modeling for automated theorem proving}.
\newblock \emph{arXiv preprint arXiv:2009.03393}.

\bibitem[{Raffel et~al.(2020)Raffel, Shazeer, Roberts, Lee, Narang, Matena,
  Zhou, Li, and Liu}]{t5}
Colin Raffel, Noam Shazeer, Adam Roberts, Katherine Lee, Sharan Narang, Michael
  Matena, Yanqi Zhou, Wei Li, and Peter~J Liu. 2020.
\newblock \href {https://doi.org/10.48550/arXiv.1910.10683} {Exploring the
  limits of transfer learning with a unified text-to-text transformer}.
\newblock \emph{Journal of Machine Learning Research}, 21:1--67.

\bibitem[{Ray(2007)}]{Ray2007}
Oliver Ray. 2007.
\newblock \href {https://doi.org/10.1007/978-3-540-71986-1_5} {Automated
  abduction in scientific discovery}.
\newblock In \emph{Model-Based Reasoning in Science, Technology, and Medicine},
  pages 103--116. Springer.

\bibitem[{Saha et~al.(2020)Saha, Ghosh, Srivastava, and Bansal}]{prover}
Swarnadeep Saha, Sayan Ghosh, Shashank Srivastava, and Mohit Bansal. 2020.
\newblock \href {https://dx.doi.org/10.18653/v1/2020.emnlp-main.9} {Prover:
  Proof generation for interpretable reasoning over rules}.
\newblock In \emph{Proceedings of the 2020 Conference on Empirical Methods in
  Natural Language Processing (EMNLP)}, pages 122--136.

\bibitem[{Saha et~al.(2021)Saha, Yadav, and
  Bansal}]{saha-etal-2021-multiprover}
Swarnadeep Saha, Prateek Yadav, and Mohit Bansal. 2021.
\newblock \href {https://doi.org/10.18653/v1/2021.naacl-main.287}
  {multi{PR}over: Generating multiple proofs for improved interpretability in
  rule reasoning}.
\newblock In \emph{Proceedings of the 2021 Conference of the North American
  Chapter of the Association for Computational Linguistics: Human Language
  Technologies}, pages 3662--3677, Online. Association for Computational
  Linguistics.

\bibitem[{Saparov and Mitchell(2021)}]{saparov2021generative}
Abulhair Saparov and Tom~M Mitchell. 2021.
\newblock \href {https://doi.org/10.48550/arXiv.2105.02486} {A generative
  symbolic model for more general natural language understanding and
  reasoning}.
\newblock \emph{arXiv preprint arXiv:2105.02486}.

\bibitem[{Tafjord et~al.(2021)Tafjord, Dalvi, and
  Clark}]{tafjord-etal-2021-proofwriter}
Oyvind Tafjord, Bhavana Dalvi, and Peter Clark. 2021.
\newblock \href {https://doi.org/10.18653/v1/2021.findings-acl.317}
  {{P}roof{W}riter: Generating implications, proofs, and abductive statements
  over natural language}.
\newblock In \emph{Findings of the Association for Computational Linguistics:
  ACL-IJCNLP 2021}, pages 3621--3634, Online. Association for Computational
  Linguistics.

\bibitem[{Vaswani et~al.(2017)Vaswani, Shazeer, Parmar, Uszkoreit, Jones,
  Gomez, Kaiser, and Polosukhin}]{vaswani2017attention}
Ashish Vaswani, Noam Shazeer, Niki Parmar, Jakob Uszkoreit, Llion Jones,
  Aidan~N Gomez, {\L}ukasz Kaiser, and Illia Polosukhin. 2017.
\newblock \href {https://doi.org/10.48550/arXiv.1706.03762} {Attention is all
  you need}.
\newblock In \emph{Advances in neural information processing systems}, pages
  5998--6008.

\bibitem[{Welleck et~al.(2021)Welleck, Liu, Bras, Hajishirzi, Choi, and
  Cho}]{welleck2021naturalproofs}
Sean Welleck, Jiacheng Liu, Ronan~Le Bras, Hannaneh Hajishirzi, Yejin Choi, and
  Kyunghyun Cho. 2021.
\newblock \href {https://doi.org/10.48550/arXiv.2104.01112} {Naturalproofs:
  Mathematical theorem proving in natural language}.
\newblock \emph{arXiv preprint arXiv:2104.01112}.

\bibitem[{Wolf et~al.(2020)Wolf, Debut, Sanh, Chaumond, Delangue, Moi, Cistac,
  Rault, Louf, Funtowicz, Davison, Shleifer, von Platen, Ma, Jernite, Plu, Xu,
  Le~Scao, Gugger, Drame, Lhoest, and Rush}]{wolf-etal-2020-transformers}
Thomas Wolf, Lysandre Debut, Victor Sanh, Julien Chaumond, Clement Delangue,
  Anthony Moi, Pierric Cistac, Tim Rault, Remi Louf, Morgan Funtowicz, Joe
  Davison, Sam Shleifer, Patrick von Platen, Clara Ma, Yacine Jernite, Julien
  Plu, Canwen Xu, Teven Le~Scao, Sylvain Gugger, Mariama Drame, Quentin Lhoest,
  and Alexander Rush. 2020.
\newblock \href {https://doi.org/10.18653/v1/2020.emnlp-demos.6} {Transformers:
  State-of-the-art natural language processing}.
\newblock In \emph{Proceedings of the 2020 Conference on Empirical Methods in
  Natural Language Processing: System Demonstrations}, pages 38--45, Online.
  Association for Computational Linguistics.

\end{thebibliography}
\bibliographystyle{acl_natbib}

\appendix

\section*{Appendices}


\section{Rephrasing method}
Table \ref{tab:rephrasing} demonstrates the method we used to rephrase rules in our more complex datasets. Our method made several binary phrasing choices to decide between 16 possible templates, providing more internal variety than PARARULE Plus but less than ParaRules.
As well as this random variation, all 3 conditions were shuffled, giving 6 possible orderings and 96 total possible rephrasings.

\begin{table*}[t]
\begin{adjustbox}{max width=\textwidth}
\begin{tabular}{|c|c|c|c|c|}
\hline
Plural? & Specific? & Also? & Then/All? & Example rephrasing \\
\hline
$\times$ & $\times$ & $\times$ & $\times$ & If something is big, is heavy, and is fierce, it is strong.\\
\hline
$\times$ & $\times$ & $\times$ & \checkmark & If something is big, is heavy, and is fierce, then it is strong.\\
\hline
$\times$ & $\times$ & \checkmark & $\times$ & If something is big, is heavy, and is fierce, it is also strong.\\
\hline
$\times$ & $\times$ & \checkmark & \checkmark & If something is big, is heavy, and is fierce, then it is also strong.\\
\hline
$\times$ & \checkmark & $\times$ & $\times$ & If an animal is big, is heavy, and is fierce, it is strong.\\
\hline
$\times$ & \checkmark & $\times$ & \checkmark & If an animal is big, is heavy, and is fierce, then it is strong.\\
\hline
$\times$ & \checkmark & \checkmark & $\times$ & If an animal is big, is heavy, and is fierce, it is also strong.\\
\hline
$\times$ & \checkmark & \checkmark & \checkmark & If an animal is big, is heavy, and is fierce, then it is also strong.\\
\hline
\checkmark & $\times$ & $\times$ & $\times$ & Things that are big, are heavy, and are fierce, are strong.\\
\hline
\checkmark & $\times$ & $\times$ & \checkmark & All things that are big, are heavy, and are fierce, are strong.\\
\hline
\checkmark & $\times$ & \checkmark & $\times$ & Things that are big, are heavy, and are fierce, are also strong.\\
\hline
\checkmark & $\times$ & \checkmark & \checkmark & All things that are big, are heavy, and are fierce, are also strong.\\
\hline
\checkmark & \checkmark & $\times$ & $\times$ & Animals that are big, are heavy, and are fierce, are strong.\\
\hline
\checkmark & \checkmark & $\times$ & \checkmark & All animals that are big, are heavy, and are fierce, are strong.\\
\hline
\checkmark & \checkmark & \checkmark & $\times$ & Animals that are big, is heavy, and is fierce, are also strong.\\
\hline
\checkmark & \checkmark & \checkmark & \checkmark & All animals that are big, are heavy, and are fierce, are also strong.\\

\hline
\end{tabular}
\end{adjustbox}
\caption{\label{tab:rephrasing}A diagram demonstrating the successive changes we make to the AbductionRules knowledge bases.}
\end{table*}
\section{Lossless errors}
The following encompass all errors we considered lossless - i.e. close enough to the ground truth answer to be reasonably counted as correct.
\begin{itemize}
\item
Unnecessary inclusion of 'the', as in "The Bob is small."
\item
Omission of 'the', as in "Cat is smart."
\item
Unnecessary inclusion of 'is', as in "The lion is attacks the mouse."
\item
Omission of 'is', as in "The squirrel funny."
\item
Inclusion of words that are never included in our answers, specifically 'and', 'are', and 'a'.
\item
Renaming the entity to better resemble training examples; for example, person-domain models sometimes replaced 'the crocodile' with 'Cro' while animal-domain models replaced 'Bob' with 'the bobster'. 
\item
Looping the correct answer or some part thereof, as in "The dog is is is is is small." or "The rabbit is rabbit is adorable."
\item
Incorrect capitalisation, as in "The anne is wealthy."
\item
Omission of spaces, as in "Thebob is small."
\end{itemize}

\section{Abduction-Person-Simple example}

Figure \ref{abductionrules} contains an example item from Abduction-Person-Simple, similarly to Figure \ref{animalsimpleexample}'s example from Abduction-Animal-Simple.

\begin{figure}[h]
\begin{adjustbox}{max width=\columnwidth}

\fbox{%
  \parbox{\columnwidth}{%
  
\textit{Context(Facts+Rules):}

\textit{Facts:}
Anne is dull. Dave is nice. Erin is tiny. \textbf{Fiona is \colorbox{pink}{\color{black}high}.} \textbf{Fiona is \colorbox{cyan}{\color{black}strong}.} Erin is small. Dave is clever. Fiona is heavy. Anne is sad. Anne is rough. Erin is thin.

\textit{Rules:}
\textbf{All things that are \colorbox{green}{\color{black}big}, are \colorbox{pink}{\color{black}high}, and are \colorbox{cyan}{\color{black}strong}, are also \colorbox{yellow}{\color{black}huge}.} If something is poor, is small, and is nice, it is also huge. All things that are high, are rough, and are little, are also smart. All things that are clever, are quiet, and are dull, are smart. People that are big, are dull, and are clever, are also short. If a person is thin, is small, and is little, that person is short. If a person is thin, is strong, and is quiet, that person is imperfect. Things that are little, are small, and are nice, are short. If a person is high, is poor, and is rough, then that person is also imperfect. All things that are thin, are big, and are strong, are also huge. If something is clever, is nice, and is quiet, then it is smart. If a person is poor, is rough, and is dull, then that person is imperfect. 

\textit{Question:} Fiona is \colorbox{yellow}{\color{black}huge}.

\textit{Label:} Fiona is \colorbox{green}{\color{black}big}.

  }%
}
\end{adjustbox}
\caption{An example observation, explanation, and corresponding context from Abduction-Person-Simple.
The model must output the explanation given the context and observation as input.
Facts and rules used to explain the observation are bolded while relevant attributes are highlighted.
}	
\label{abductionrules}
\end{figure}

\end{document}